\documentclass[letterpaper]{article} 
\usepackage{aaai25}  
\usepackage{times}  
\usepackage{helvet}  
\usepackage{courier}  
\usepackage[hyphens]{url}  
\usepackage{graphicx} 
\usepackage{natbib}  
\usepackage{caption} 
\usepackage{algorithm}
\usepackage{algorithmic}
\usepackage{amsfonts}
\usepackage{bm}
\usepackage{amsmath}
\usepackage{mathtools}
\usepackage{amsfonts}
\usepackage{booktabs}
\usepackage{multirow}
\usepackage{threeparttable} 
\usepackage{xcolor}
\usepackage{newfloat}
\usepackage{listings}
\DeclareCaptionStyle{ruled}{labelfont=normalfont,labelsep=colon,strut=off} 
\floatstyle{ruled}
\newfloat{listing}{tb}{lst}{}
\floatname{listing}{Listing}

\title{SrSv: Integrating Sequential Rollouts with Sequential Value Estimation for Multi-agent Reinforcement Learning}
\author {
    Xu Wan\textsuperscript{\rm 1,3},
    Chao Yang\textsuperscript{\rm 3},
    Cheng Yang\textsuperscript{\rm 3},
    Jie Song\textsuperscript{\rm 2},
    Mingyang Sun\footnote{Corresponding Author}\textsuperscript{\rm 2}
}
\affiliations {
    \textsuperscript{\rm 1}Zhejiang University \\
    \textsuperscript{\rm 2}Peking University \\
    \textsuperscript{\rm 3}Alibaba DAMO Acedemy \\
    wanxu@zju.edu.cn, xiuxin.yc@alibaba-inc.com, charis.yangc@alibaba-inc.com, jie.song@pku.edu.cn, smy@pku.edu.cn
}

\usepackage{bibentry}

\begin{document}

\maketitle

\begin{abstract}
Although multi-agent reinforcement learning (MARL) has shown its success across diverse domains, extending its application to large-scale real-world systems still faces significant challenges. Primarily, the high complexity of real-world environments exacerbates the credit assignment problem, substantially reducing training efficiency. Moreover, the variability of agent populations in large-scale scenarios necessitates scalable decision-making mechanisms.
To address these challenges, we propose a novel framework: Sequential rollout with Sequential value estimation (SrSv). This framework aims to capture agent interdependence and provide a scalable solution for cooperative MARL. Specifically, SrSv leverages the autoregressive property of the Transformer model to handle varying populations through sequential action rollout. Furthermore, to capture the interdependence of policy distributions and value functions among multiple agents, we introduce an innovative sequential value estimation methodology and integrates the value approximation into an attention-based sequential model.
We evaluate SrSv on three benchmarks: Multi-Agent MuJoCo, StarCraft Multi-Agent Challenge, and DubinsCars. Experimental results demonstrate that SrSv significantly outperforms baseline methods in terms of training efficiency without compromising convergence performance. Moreover, when implemented in a large-scale DubinsCar system with 1,024 agents, our framework surpasses existing benchmarks, highlighting the excellent scalability of SrSv.
\end{abstract}

\section{Introduction}
While multi-agent reinforcement learning (MARL) has demonstrated its feasibility in several decision-making domains such as games, robotic planning, and simulated industrial control \cite{kober2013reinforcement, vinyals2019grandmaster,andrychowicz2020learning, wang2022stabilizing}, applying it to large-scale real-world systems remains an open challenge. 
The interaction complexity among a large population of agents creates a fundamental difficulty in identifying individual agents' contributions to the global reward signal, which significantly exacerbates the credit assignment problem of MARL \cite{foerster2018counterfactual}. Most current MARL algorithms adopt the centralized training with decentralized execution (CTDE) paradigm to partially alleviate this issue \cite{lowe2017multi}, where the CTDE framework allows for a straightforward extension of single-agent policy gradient theorems by providing agents with access to global information and other agents' observations during training.

Within the CTDE framework, existing popular MARL algorithms such as QMIX \cite{rashid2018qmix}  and Quality-Diversity Policy Pursuit (Q-DPP) \cite{yang2020q} typically using value decomposition theory to represent the global Q-value as an aggregation of individual agent values, lacking effective modeling capabilities for intricate multi-agent interactions. Despite achieving successes on small-scale tasks, these value decomposition-based methods face limitations in scaling to complex multi-agent scenarios with high-dimensional joint policy spaces \cite{wang2020qplex}.

To circumvent the restrictive assumptions on the decomposability of the joint value function, the multi-agent advantage decomposition theory and sequential policy update algorithms have been proposed \cite{kuba2021settling, zhang2021fop}. The advantage decomposition lemma aims to characterize an agent's advantage increment over preceding agents' decisions, providing insights into the emergence of cooperative behavior through a sequential decision-making process. 

Building upon this theory, the advantage decomposition-based methods such as Heterogeneous-Agent PPO (HAPPO) and Heterogeneous-Agent Soft Actor-Critic (HASAC) \cite{kuba2022trust, liu2024maximum} have been developed to enable policy iteration with monotonic improvement guarantees. Meanwhile, unlike the sequential policy updates, some researchers, exemplified by the Multi-Agent Transformer (MAT) and Action-dependent Q-learning (ACE) \cite{wen2022multi, li2023ace}, capture cooperative intentions among multiple agents from the perspective of multi-agent action rollouts. They leverage the sequence model to generate actions agent by agent. By incorporating the multi-agent advantage decomposition theorem into an encoder-decoder architecture, it realizes sequential action generation with a monotonic performance improvement guarantee.
 
Nonetheless, a potential limitation of MAT lies in that it estimates solely the joint advantage function, lacking the fine-grained estimation of individual agents' advantage functions compared to individual advantage-based approaches. As the scale of the system increases, the joint estimation process will become more sample-inefficient and computationally demanding. 

While prior work has proposed sequential policy update algorithms, like preceding-agent off-policy correction (PreOPC) \cite{wang2023order}  and agent-by-agent policy optimization (A2PO) \cite{wang2023order} to estimate individual advantage functions, retaining monotonic improvement guarantees on both the joint policy and each agent's policy, determining the optimal decision-making order among agents can be computationally intractable, especially in large-scale systems. Moreover, the "all agents at once" rollout strategy of PreOPC and A2PO, also neglects the sequential correlations inherent in multi-agent decision-making processes.

To address these challenges, we propose a novel paradigm that synergizes the individual-centric value estimation from sequential update methods with the sequential rollout strategy, termed \textbf{SrSv} (\textbf{S}equential \textbf{r}ollouts with \textbf{S}equential \textbf{v}alue estimation). 
The SrSv paradigm leverages the inherent autoregressive property of Transformer models, making it possible to effectively capture inter-agent correlations during the action rollout process. Simultaneously, by estimating individual-centric value functions sequentially, SrSv significantly enhances the training efficiency of MARL, especially in dealing with complex tasks.

Experimental results on three benchmarks - Multi-Agent MuJoCo (MAMuJoCo) \cite{de2020deep}, StarCraft Multi-Agent Challenge (SMAC) \cite{samvelyan2019starcraft}, and DubinsCar \cite{zhang2023neural} - demonstrate that SrSv can outperform strong baselines in terms of convergence speed without sacrificing training performance. Furthermore, scaling SrSv to a large-scale DubinsCar system with 1024 agents further evidences its excellent scalability.

\section{Related Work}

The CTDE paradigm has been widely adopted in the MARL community. Early works such as VDN \cite{sunehag2017value} and QMIX \cite{rashid2018qmix} proposed value decomposition methods satisfying the Individual-Global-Max (IGM) condition. Building upon this foundation, more advanced approaches like QTRAN \cite{son2019qtran} and QPLEX \cite{wang2020qplex}, aimed to relax the IGM constraints while maintaining factorized value functions. However, these methods often struggle with complex multi-agent tasks due to their restrictive assumptions and limited expressiveness in capturing inter-agent dependencies.

To address these limitations,  recent research has introduced advantage decomposition theory, eliminating the need for any assumptions on agents sharing parameters or the joint value function being decomposable. This breakthrough has led to the development of two parallel frameworks: the sequential update scheme and the sequential rollout scheme. 

The sequential update scheme, represented by methods such as HAPPO \cite{kuba2021trust} and A2PO \cite{wang2023order}, leverages individual advantage estimations to improve training efficiency and stability in multi-agent settings. In particular, HAPPO introduces a trust region method for multi-agent policy optimization, ensuring monotonic improvement of the joint policy. A2PO further enhances this approach to improve the sample efficiency and addresses non-stationarity issues in their theory.  However, these methods may struggle with scalability in environments with a large number of agents due to the high computational complexity during sequential policy updates. Simultaneously, the sequential rollout scheme, exemplified by MAT \cite{wen2022multi}, utilizes a Transformer-based architecture to model the sequential decision-making process, enabling more effective coordination among agents. Experiments on both SMAC and multi-agent MuJoCo tasks have shown that MAT can effectively transfer knowledge from simpler tasks to more complex and diverse scenarios, showcasing the potential of Transformer-based models in the scalability of MARL. Nevertheless, in MAT, each agent’s policy is updated according to the joint advantage function, which may lead to overall low training efficiency, especially in complex cooperative scenarios.

\section{Preliminaries}
\subsection{Cooperative MARL Problems Formulation}
We consider formulating the cooperative MARL problems as a Dec-POMDP \cite{bernstein2002complexity}, which can be described by a tuple $\langle N, S, \bm{\mathcal{A}}, \bm{\mathcal{O}}, P, R, \gamma \rangle$. Here, $N = \{1, ..., n\}$ represents the set of $n$ agents. $S$ denotes the global state space, $\bm{\mathcal{A}} = \prod\nolimits_i {\mathcal{A}_i}$ and $\bm{\mathcal{O}} = \prod\nolimits_i {\mathcal{O}_i}$ represent the joint action and observation spaces, respectively, where $\mathcal{A}_i$ and $\mathcal{O}_i$ are the individual action and observation spaces for agent $i$. The transition function $P: S \times \bm{\mathcal{A}} \times S \rightarrow [0, 1]$ determines the probability of transitioning to a new state given the current state and joint action. $R: S \times \bm{\mathcal{A}} \rightarrow [-R_{\text{max}}, R_{\text{max}}]$ is the bounded global reward function, shared by all agents, and $\gamma \in [0, 1)$ is the discount factor. 

At time step $t$, every agent $i \in N$ receives a local observation $o^i_t \in \mathcal{O}_i$ from the global state $s_t$ and selects an action $a^i_t$ based on its policy $\pi^i$. The joint action $\bm{a}_t = (a^1_t, ..., a^n_t)$ leads to a new state $s_{t+1}$ according to $P$, and a team reward $r_t = R(s_t, \bm{a}_t)$. We define the joint policy $\pi(\bm{a}_t|\bm{o}_t)$ as a conditional probability of the joint action $\bm{a}_t$ given all the agents' observations $\bm{o}_t = (o^1_t, ..., o^n_t)$. The objective in cooperative MARL is to find an optimal joint policy $\pi^*$ that maximizes the expected cumulative discounted reward:

\begin{equation}
\pi^* = \arg\max_\pi \mathbb{E}_\pi \left[\sum_{t=0}^{\infty} \gamma^t R(s_t, \bm{a}_t)\right]
\end{equation}

\subsection{Advantage Decomposition Theorem}

The advantage decomposition theory decomposes the joint advantage function into individual agent advantages, allowing for more efficient and effective multi-agent learning. The key insight is that the joint advantage function can be expressed as:
\begin{equation}
A(\bm{o}_t, \bm{a}_t) = \sum_{i=1}^n A_t^i(\bm{o}_t, a_t^i|\bm{a}_t^{1:i-1})
\end{equation}
where $A(\bm{o}_t, \bm{a}_t)$ is the joint advantage function in time step $t$, $A_i(\bm{o}_t, a^i|\bm{a}_t^{1:i-1})$ is the advantage function for agent $i$ conditioned on the actions of preceding
agents, and $\bm{a}_t^{1:i-1} = (a_t^1, ..., a_t^{i-1})$  represents the actions of agents with decision order lower than $i$.

This decomposition allows for the factorization of the joint policy into a product of individual policies:
\begin{equation}
\pi(\bm{a}_t|\bm{o}_t) = \prod_{i=1}^n \pi^i(a_t^i|\bm{o}_t, \bm{a}_t^{1:i-1})
\end{equation}
where $\pi^i(a_t^i|\bm{o}_t, \bm{a}_t^{1:i-1})$ is the policy of agent $i$ at time step $t$, conditioned on the joint observation and the actions of agents with lower decision order.

\subsection{Multi-Agent Transformer}
Based on the decomposition of advantage, MAT treats the MARL process as a sequence of tokens and uses a transformer architecture to model the dependencies among multi-agents decision-making. 

Specifically, the sequence of input tokens for each time step $t$ is represented as $[o_t^1, \ldots, o_t^n]$. MAT uses a transformer encoder to process the sequence of tokens and generate observation embeddings as $[{\widehat{\bm{o}_t^1}}, \ldots, {\widehat{\bm{o}_t^n}}]$ and $n$ value functions as $[V({\widehat{\bm{o}_t^1}}), \ldots, V({\widehat{\bm{o}_t^n}})]$. Note that ${\widehat{\bm{o}_t^i}}$ capture both the information specific to agent $i$ and the high-level interrelationships that represent agents' interactions. The decoder then autoregressively generates actions for each agent. For the $i$-th agent, the decoder takes the concatenation of the observation embeddings ${\widehat{\bm{o}_t^{1:i}}} = [{\widehat{\bm{o}_t^1}}, \ldots, {\widehat{\bm{o}_t^i}}]$ and the preceding generated actions $[\bm{a}_t^{1:i-1}]$ as input. The output of the decoder is the policy $\pi^i(a_t^i|{\widehat{\bm{o}_t^{1:i}}}, \bm{a}_t^{1:i-1})$. In the actual code implementation, the transformer decoder estimates the joint policy distribution $\pi(a_t^i|{\widehat{\bm{o}_t^{1:i}}}, \bm{a}_t^{1:i-1})$ for all agents and only the $i$-th agent's policy is used.

\begin{figure}[t]
  \centering
  \includegraphics[width=\columnwidth]{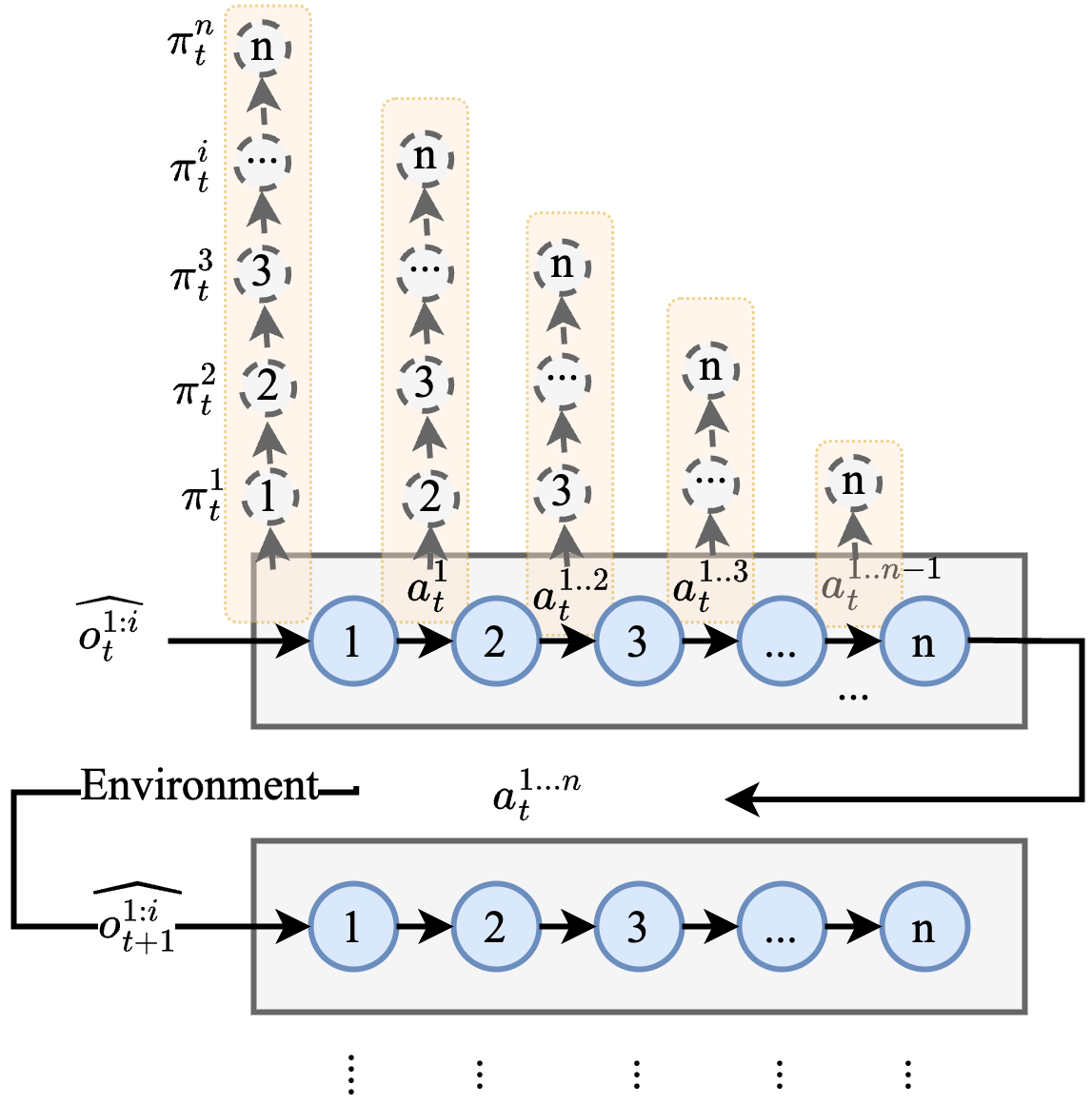} 
  \caption{The modeling of agent interdependence for sequential decision-making.}
  \label{fig:motivation}
\end{figure}

\section{Methodology}
\begin{figure*}[t]
  \centering
  \includegraphics[width=2.1\columnwidth]{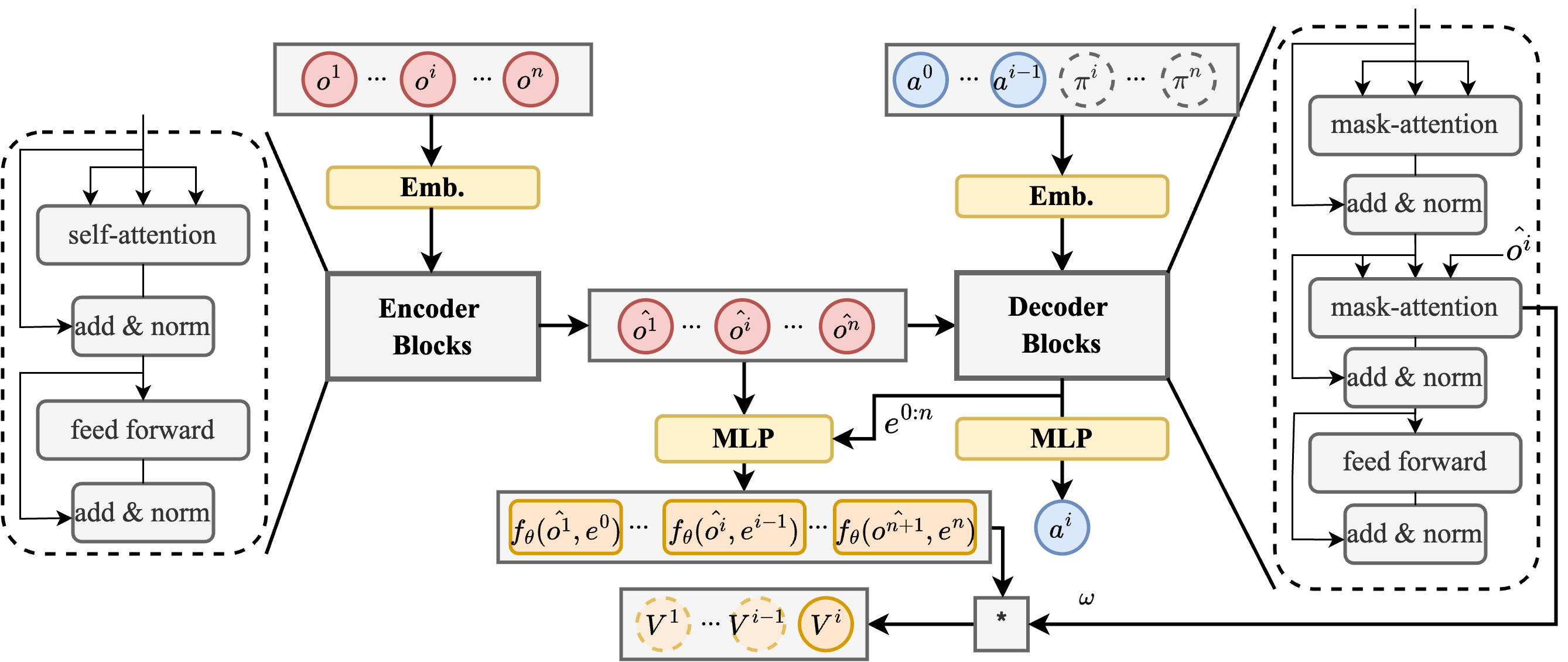} 
  \caption{The encoder-decoder architecture of SrSv.}
  \label{fig:framework}
\end{figure*}

In this section, we first introduce the modeling of agent interdependence when multiple agents follow the paradigm of sequential decision-making, as shown in Fig \ref{fig:motivation}. Based on the modeling of agent interdependence, we can obtain an accurate description of policy distribution and value function. Then, we present the details of SrSv, including the transformer-based neural network architecture and the corresponding algorithm for training neural networks.

\subsection{Modeling of Agent Interdependence for Sequential Decision-making}
Here, we mainly focus on sequential decision-making scenarios where all agents make decisions in sequence the current agent can access its predecessors’ behaviors, and then takes its optimal decision. Such a decision-making paradigm can better utilize the interdependence between agent strategies to achieve better decisions. In this way, for the $i$-th agent at time $t$, its action can be recursively expressed as:
\begin{subequations}
	\begin{gather}
		a_t^1\sim{\pi_t ^1}\left( \cdot | {\widehat{\bm{o}_t^{1}}} \right)\\
		a_t^i\sim{\pi_t ^i}\left( \cdot |  {{\widehat{\bm{o}_t^{1:i}}},\bm{a}_t^{1:i - 1}} \right), i = 2, \cdots ,n
	\end{gather}
\end{subequations}
where $\pi_t^i$ is the policy distribution of agent $i$ in time step $t$, ${\widehat{\bm{o}_t^{1:n}}}$ is the global observation embeddings. The decision of the following agent depends on the decisions of its previous agents.

Each agent aims to maximize its expected return, namely the value function, by optimizing its policy distribution given its predecessors' strategies. For the value function of agent $i$ at time $t$, there are also temporal dependence characteristics. Here, the $i$-th value function $V^i$ is not only related to the observation embedding ${\widehat{\bm{o}_t^{i}}}$, but also related to the predecessors' behavior and its successor's policy distribution as follows:
\begin{subequations}
	\label{value_function}
	\begin{gather}
		V^1\left( {\widehat{\bm{o}_t^{1}}} \right) =  \;\;\;\mathop {\mathbb{E}}\limits_{\mathclap{\bm{a}_t^{1:n}\sim\pi _t^{1:n}}} \;\;\;\; \left[ {{V^{\pi _t^{1:n}}}\left( {{\widehat{\bm{o}_t^{1}}},\bm{a}_t^{1:n}} \right)} \right]\\
        \begin{gathered}
    		V^i\left( {{\widehat{\bm{o}_t^{i}}},\bm{a}_t^{1:i - 1}} \right) = \;\;\; \mathop {\mathbb{E}} \limits_{\mathclap{\bm{a}_t^{i:n}\sim\pi _t^{i:n}}} \;\;\;\; \left[ {{V^{\pi _t^{i:n}}}\left( {{\widehat{\bm{o}_t^{i}}},\bm{a}_t^{1:i - 1},\bm{a}_t^{i:n}} \right)} \right],\\
            i = 2, ..., n
        \end{gathered}
	\end{gather}
\end{subequations}
where $V^{\pi _t^{i:n}}$ refers to the global action value based on the policies $\pi _t^{i:n}$ after the preceding $i-1$ agents have made their decisions. Note that the physical meaning of $V^{\pi _t^{i:n}}$ is more closely related to $Q^{\pi _t^{i:n}}$ in traditional RL definition. For the sake of simplicity in our notation, we use $V^{\pi _t^{i:n}}$ here.

In order to capture the interdependence of policy distribution and value functions between different agents, in the next section, we design an attention-based neural network architecture for approximating each agent's value function and making decisions sequentially.

\subsection{Attention-based Neural Network for Value Approximation and Policy Optimization}
We use an encoder-decoder-based neural network architecture like Transformer so that agents can make decisions in an auto-regressive way. The framework of SrSv is shown in Fig. \ref{fig:framework}. The pseudo-code of SrSv is listed in Alg. \ref{alg: SrSv}.

\begin{algorithm}[tb]
\caption{\textbf{S}equential \textbf{r}ollout with \textbf{S}equential \textbf{v}alue estimation (SrSv)}
\label{alg: SrSv}
\begin{algorithmic}[1]
\STATE \textbf{Input:} Batch size $B$, number of agents $n$, number of episodes $K$, max steps per episode $T$
\STATE \textbf{Initialize:} The encoder $\{ \Phi \}$, decoder $\{ \theta_d \}$ including MLP $\theta$ for value estimation, the replay buffer $\mathcal{B}$. \\
\COMMENT {// \textit{Inference Phase}}
\FOR{$ = 0$ to $K-1$}
    \FOR{$t = 0$ to $T-1$}
        \STATE Collect $o_t^1, o_t^2, ..., o_t^n$ from environments. 
        \STATE Output the representation sequence ${\widehat{\bm{o}_t^{1}}}, {\widehat{\bm{o}_t^{1}}}, ..., {\widehat{\bm{o}_t^{n}}}$ by feeding embedded observations to the encoder.
        \FOR{$i = 0$ to $n$}
            \STATE Generate $a_t^{i}$ with the decoder based on ${\widehat{\bm{o}_t^{1:n}}}$ in environments and $\bm{a}_t^{1:i-1}$.
        \ENDFOR
        \STATE Execute joint actions $\bm{a}_t^{1:n}$ and collect reward $r_t$.
        \STATE Insert $(\bm{o}_t^{1:n}, \bm{a}_t^{1:n}, r_t)$ into $\mathcal{B}$.
    \ENDFOR \\
    \COMMENT {// \textit{Training Phase}}
    \STATE Random sample a batch of $B$ transitions from $\mathcal{B}$.
    \FOR{$i = 0$ to $n$}
        \STATE Calculate $V^{\pi _t^{i:n}}(\widehat{\bm{o}_{t}^i}, \bm{a}_t^{1:i-1}, \underset{\bm{a}_t^{i:n}}{\operatorname{argmax}} \ \pi_t^{i:n})$ using  attention matrix $\bm{w}$ and the decoder block's output $e$ as the $i$-th value estimation $V^i\left( {{\widehat{\bm{o}^{i}}},\bm{a}^{1:i - 1}} \right)$ 
        \STATE Compute the joint advantage function $A_t^i$ using GAE by Eq. (\ref{eq:gae}).
    \ENDFOR
    \STATE Update the encoder and decoder by minimizing $L_{V} + L_{\pi}$ with gradient descent.
\ENDFOR
\end{algorithmic}
\end{algorithm}

\paragraph{Encoder.}In the encoder network, we adopt the popular architecture similar to the Transformer framework. Each block in the encoder consists of a self-attention mechanism and a multi-layer perceptron (MLP), complemented by residual connections. These residual connections serve to mitigate gradient vanishing and prevent network degradation as the depth increases, thereby facilitating the extraction of robust observation embeddings for each agent. Different from MAT, here we remove the MLP for value function calculation in the encoder. This modification is motivated by the understanding that each agent's value function is not solely dependent on its observation embedding. As demonstrated in Eq. ({\ref{value_function}}), the value function should also account for the actions of preceding agents and the policy distributions of succeeding agents.

\paragraph{Decoder.}The decoder network retains the sequence of decoding blocks with masked attention for maintaining the auto-regressive property of the decision-making process. Besides, we employ a shared-parameter MLP, denoted as $f_\theta$, across all agents after the decoder blocks. For each agent $i$, the input to $f_\theta$ is the observation embedding $\widehat{\bm{o}_{t}^i}$ from encoder and $(i-1)$-th agent's  embedding obtained through auto-regressive decoding, denoted as $e^{i-1}$. The output of $f_\theta$ can be interpreted as an individual value function estimate for the $i$-th agent. 

To capture the interdependencies between agents and obtain the estimate of $V^{\pi _t^{i:n}}$ in Eq. (\ref{value_function}), we utilize the decoder's attention matrix $\bm{w}$ to weight each agent's MLP-derived value function. Crucially, the use of masked attention ensures that we adhere to the assumptions of advantage decomposition, as each agent can only consider the actions of preceding agents in the decision sequence. 
Similar in MAT, we introduce an arbitrary symbol $a^0$ to indicate the start of decoding, allowing us to transform $V^1(\widehat{\bm{o}_{t}^1})$ into $V^1(\widehat{\bm{o}_{t}^1}, a^0)$. 
Besides, $V^{\pi _t^{i:n}}$ is estimated by using the last row of the attention matrix to weight the individual $f_{\theta}$ as follows:
\begin{subequations}
\label{eq:v_pi}
\begin{gather}
V^{\pi _t^{i:n}}(\widehat{\bm{o}_{t}^i}, \bm{a}_t^{1:i-1}, \bm{a}_t^{i:n}) = \sum_{j=1}^{\textcolor{blue}{n+1}} w_{n+1,j} \cdot f_\theta(\widehat{\bm{o}_{t}^j}, e^{j-1})\\
\widehat{\bm{o}_{t}^{n+1}} = \widehat{\bm{o}_{t}^{i}}, \ i = 1, ..., n
\end{gather}
\end{subequations}
where $w_{n+1,j}$ is the attention weight from the $(n+1)$-th row of the attention matrix $\bm{w}$, corresponding to the $j$-th agent.

To simplify the calculation of $V^i$, we approximate the expected form of $V^{\pi _t^{i:n}}(\widehat{\bm{o}_{t}^i}, \bm{a}_t^{1:i-1}, \bm{a}_t^{i:n})$ over $\bm{a}_t^{i:n} \sim \pi_t^{i:n}$ with $V^{\pi _t^{i:n}}(\widehat{\bm{o}_{t}^i}, \bm{a}_t^{1:i-1}, \underset{\bm{a}_t^{i:n}}{\operatorname{argmax}} \ \pi_t^{i:n})$ in the code implementation.

In the training phase of neural networks, Proximal Policy
Optimization (PPO) \cite{schulman2017proximal} is used to update neural network parameters for value function approximation and policy optimization. For the value function approximation, we minimize the sum of Bellman errors of all agents as follows:
\begin{equation}
	L_v = \sum\limits_{i = 1}^n {\sum\limits_{t = 0}^{T - 1} {\left( \begin{array}{l}
				R\left( {{s_t},\bm{a}_t^{1:n}} \right) + \gamma V^i\left( {\widehat{\bm{o}_{t + 1}^i},\bm{a}_{t + 1}^{1:i - 1}} \right)\\
				- {V^i}\left( {\widehat{\bm{o}_{t}^i},\bm{a}_{t}^{1:i - 1}} \right)
			\end{array} \right)} }
\end{equation}

As to policy optimization, we maximize the following clipping PPO objective and use policy gradient to update the neural network parameters:
\begin{equation}
	L_{\pi} = \sum\limits_{i = 1}^n {\sum\limits_{t = 0}^{T - 1} {\min \left( {\alpha _t^iA_t^i,{\rm{clip}}\left( {\alpha _t^i,1 \pm \epsilon } \right)A_t^i} \right)} }
\end{equation}
where $\alpha_t^{i} =\frac{\pi (a_t^i|\widehat{\bm{o}_{t}^{1:i}}, \bm{a}_{t}^{1:i-1})}{\pi_{{\rm{old}}}(a_t^i|\widehat{\bm{o}_{t}^{1:i}}, \bm{a}_{t}^{1:i-1})}$ is the probability ratio. The advantage of the $i$-th agent is approximated using Generalized Advantage Estimation (GAE) \cite{schulman2015high} as follows:
\begin{equation}
	A_t^i = \sum\limits_{j = 0}^h {{{\left( {\gamma \lambda } \right)}^j}\left( \begin{array}{l}
			R\left( {{s_t},a_t^{1:n}} \right) + \gamma V^i\left( {\widehat{\bm{o}_{t + 1}^i},\bm{a}_{t + 1}^{1:i - 1}} \right)\\
			- {V^i}\left( {\widehat{\bm{o}_{t}^i},\bm{a}_{t}^{1:i - 1}} \right)
		\end{array} \right)}
\label{eq:gae}
\end{equation}
where $h$ is the step length of GAE.

\section{Experiments and Results}
\begin{figure*}[t]
  \centering
  \includegraphics[width=2\columnwidth]{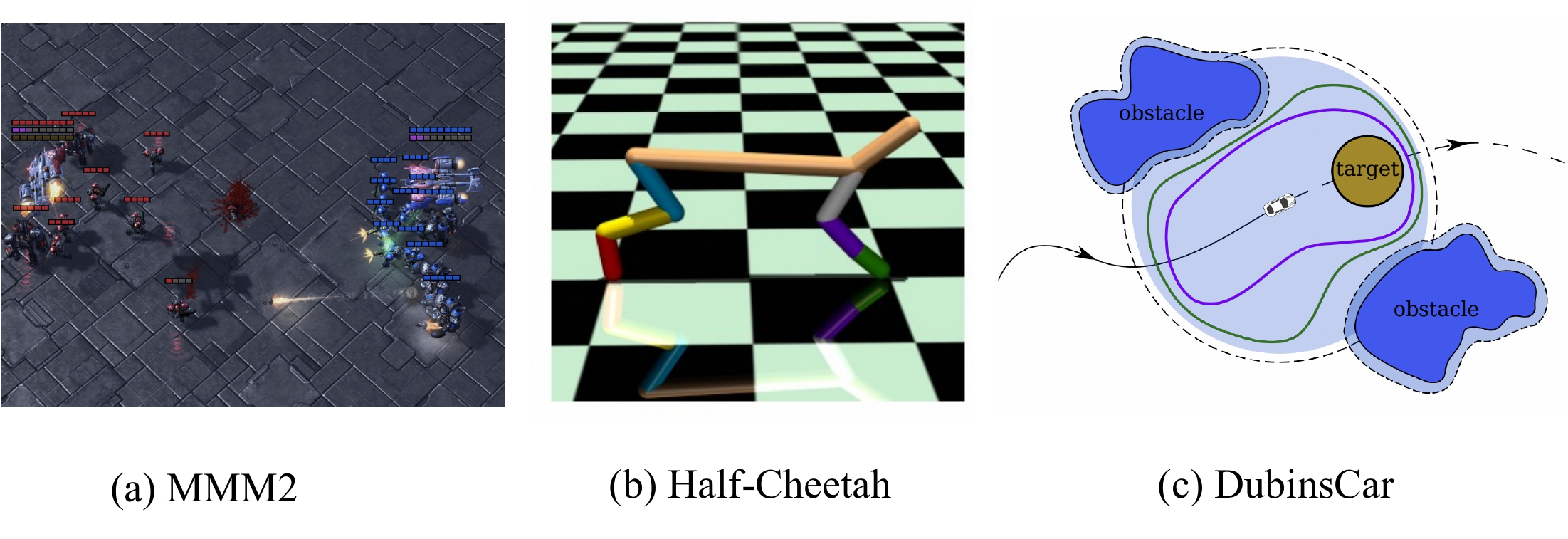} 
  \caption{Demonstrations of the multi-agent benchmarks: MMM3 in SMAC, Half-Cheetah in MA-MuJoCo and DubinsCar.}
  \label{fig:env}
\end{figure*}
\begin{figure*}[t]
  \centering
  \includegraphics[width=2.1\columnwidth]{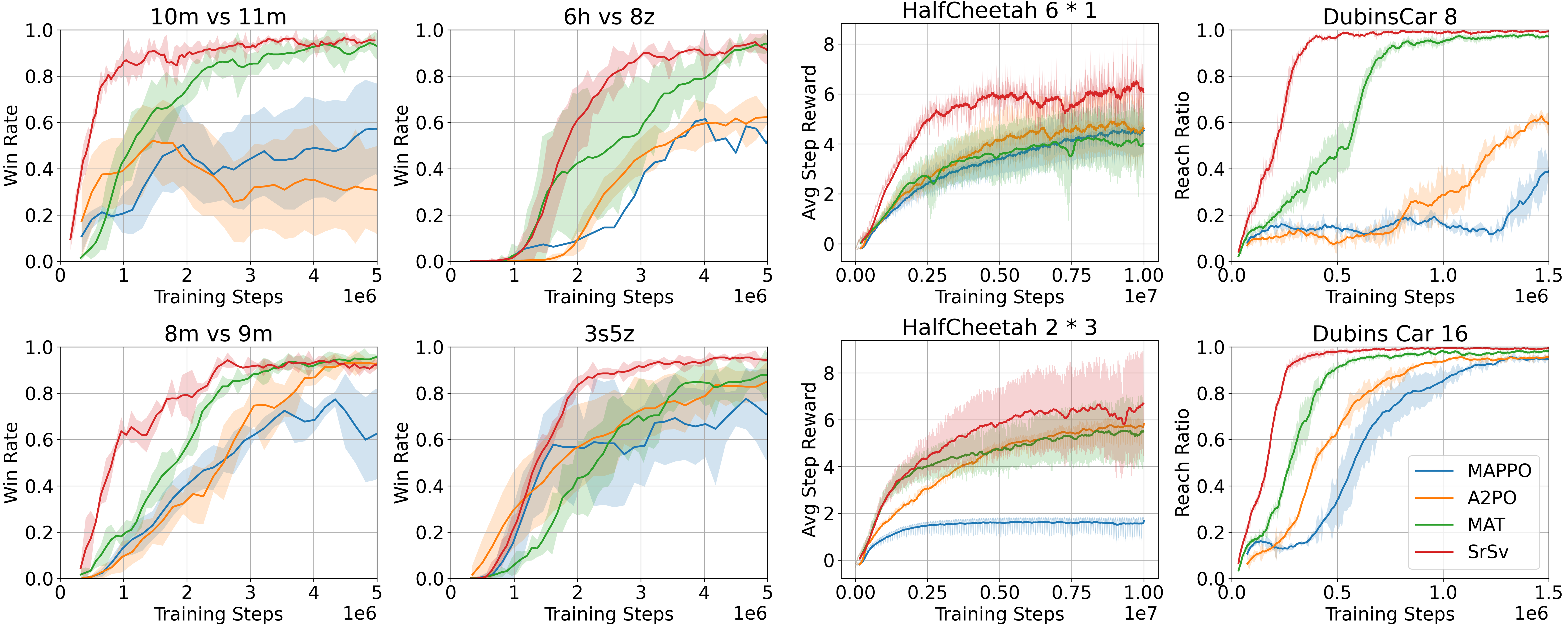} 
  \caption{Performance comparisons on cooperative MARL benchmarks among SrSv and other baselines.}
  \label{fig:training_performance}
\end{figure*}

\begin{figure}[ht]
  \centering
  \includegraphics[width=\columnwidth]{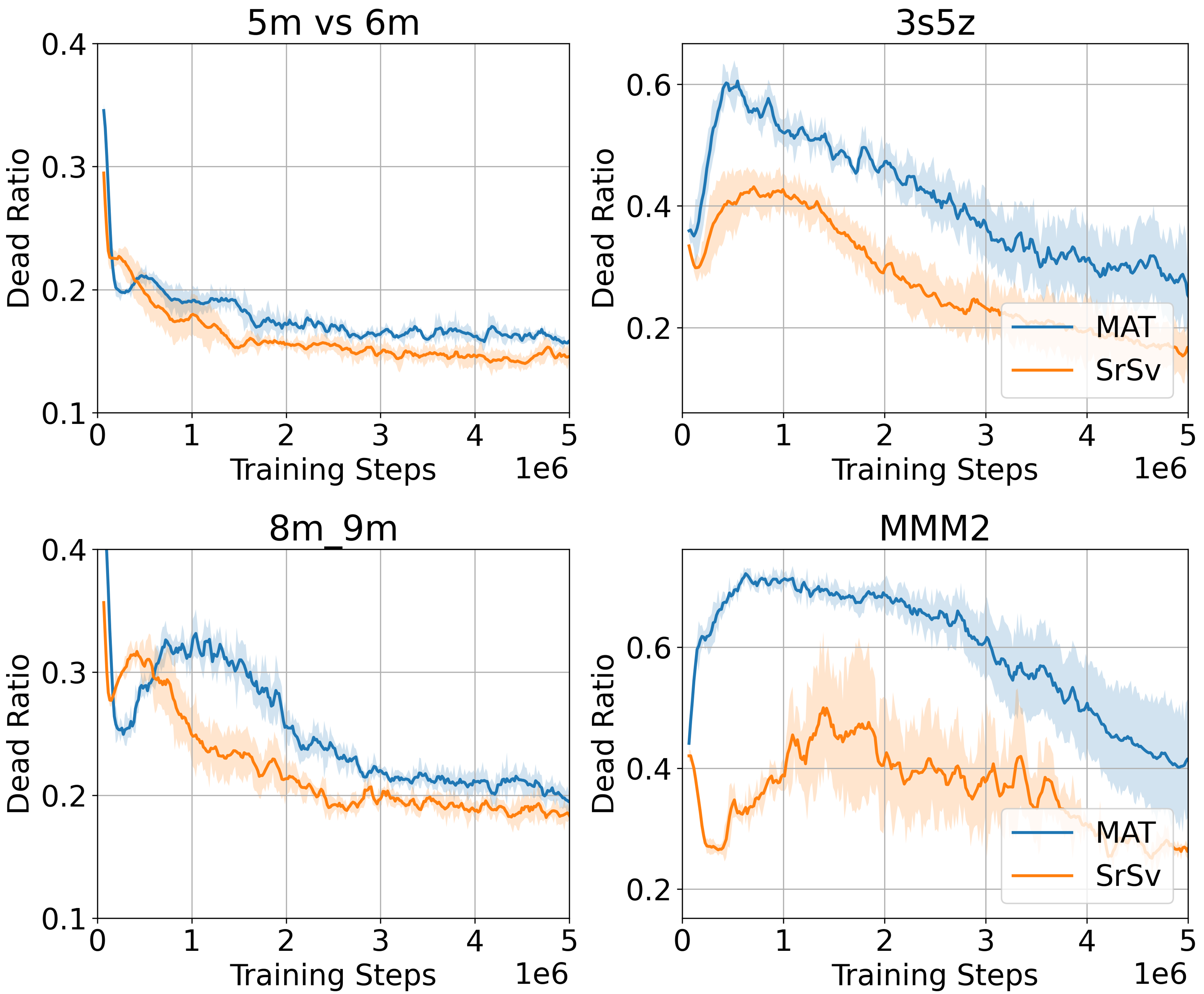} 
  \caption{Dead ratio metric comparisons on SMAC benchmark between SrSv and MAT.}
  \label{fig:training_dead_ratio}
\end{figure}

\begin{figure}[ht]
  \centering
  \includegraphics[width=\columnwidth]{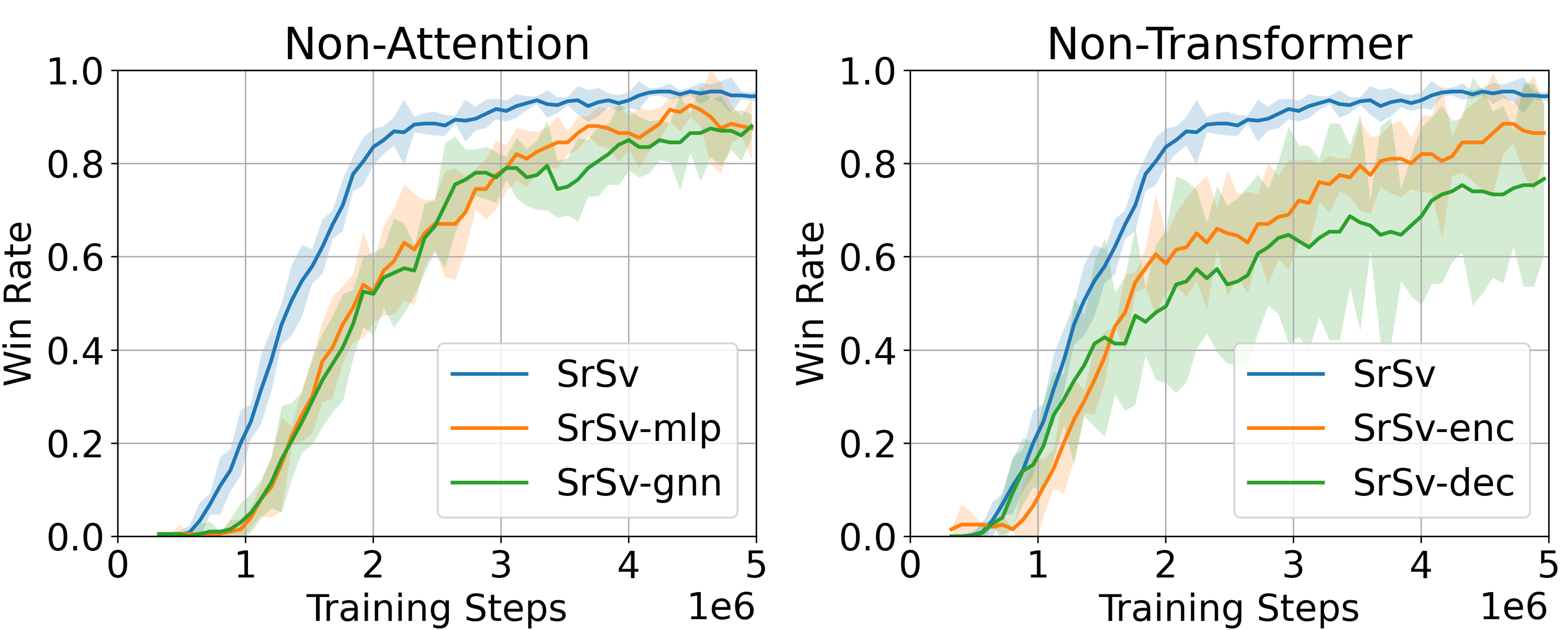} 
  \caption{Win rate comparison of different architectures of SrSv on $\texttt{3s5z}$ scenario over 4 seeds.}
  \label{fig:architecture_ablation}
\end{figure}

\begin{table*}[t]
\centering
\small
\begin{tabular}{@{}c|cccc|cccc@{}}
\toprule
\multirow{2}{*}{\begin{tabular}[c]{@{}c@{}}Eval\\      Populations\end{tabular}} & \multicolumn{4}{c|}{Train \textbf{100} epoch with 8 agents} & \multicolumn{4}{c}{Train \textbf{300} epoch with 8 agents} \\
 & MAPPO & A2PO & \multicolumn{1}{c}{MAT} & \multicolumn{1}{c|}{SrSv} & MAPPO & A2PO & \multicolumn{1}{c}{MAT} & \multicolumn{1}{c}{SrSv} \\ \midrule
8 agents & $0.33_{\small{({\pm 0.027})}}$ & $0.25_{\small{({\pm 0.019})}}$ & $0.78_{\small{({\pm 0.001})}}$ & $\textbf{0.82}_{\small{({\pm 0.029})}}$ & $0.38_{\small{({\pm 0.019})}}$ & $0.16_{\small{({\pm 0.016})}}$ & $\textbf{0.93}_{\small{({\pm 0.004})}}$ & $\textbf{0.93}_{\small{({\pm 0.005})}}$ \\
64 agents & $0.06_{\small{({\pm 0.000})}}$ & $0.08_{\small{({\pm 0.001})}}$ & $0.56_{\small{({\pm 0.002})}}$ & $\textbf{0.63}_{\small{({\pm 0.006})}}$ & $0.07_{\small{({\pm 0.001})}}$  & $0.04_{\small{({\pm 0.000})}}$ & $0.86_{\small{({\pm 0.001})}}$  & $\textbf{0.91}_{\small{({\pm 0.001})}}$ \\
256 agents & $0.06_{\small{({\pm 0.000})}}$ & $0.06_{\small{({\pm 0.000})}}$ & $0.57_{\small{({\pm 0.002})}}$ & $\textbf{0.66}_{\small{({\pm 0.002})}}$ & $0.07_{\small{({\pm 0.000})}}$ & $0.04_{\small{({\pm 0.000})}}$ & $0.87_{\small{({\pm 0.001})}}$ & $\textbf{0.92}_{\small{({\pm 0.000})}}$ \\
512 agents & $0.03_{\small{({\pm 0.000})}}$ & $0.02_{\small{({\pm 0.000})}}$ & $0.25_{\small{({\pm 0.000})}}$ & $\textbf{0.34}_{\small{({\pm 0.000})}}$ & $0.02_{\small{({\pm 0.000})}}$ & $0.03_{\small{({\pm 0.000})}}$ & $0.68_{\small{({\pm 0.000})}}$ & $\textbf{0.81}_{\small{({\pm 0.000})}}$ \\ \bottomrule
\end{tabular}
\caption{Scalability test on the DubinsCar with an increasing number of agents in the early and mid-stages of training. Each value is reported as the mean ± standard deviation of reach ratio over 10 episodes and 10 seeds.}
\label{tab:train_100_300}
\end{table*}

\begin{table*}[ht]
\centering
\small
\label{tab:full_population}
\begin{tabular}{@{}c|cccc|cccc@{}}
\toprule
\multirow{2}{*}{\begin{tabular}[c]{@{}c@{}}Eval\\      Populations\end{tabular}} & \multicolumn{4}{c|}{Train 100 epoch with \textbf{16} agents} & \multicolumn{4}{c}{Train 100 epoch with \textbf{32} agents} \\
 & MAPPO & A2PO & MAT & SrSv & MAPPO & A2PO & MAT & SrSv \\ \midrule
8 agents &  $0.18_{\small{({\pm 0.035})}}$ & $0.11_{\small{({\pm 0.011})}}$ & $\textbf{0.79}_{\small{({\pm 0.013})}}$ & $0.76_{\small{({\pm 0.019})}}$  & $0.18_{\small{({\pm 0.035})}}$ & $0.30_{\small{({\pm 0.022})}}$ & $\textbf{0.85}_{\small{({\pm 0.004})}}$ & $0.80_{\small{({\pm 0.008})}}$  \\
64agents & $0.07_{\small{({\pm 0.000})}}$ & $0.03_{\small{({\pm 0.011})}}$ & $\textbf{0.49}_{\small{({\pm 0.005})}}$ & $0.48_{\small{({\pm 0.004})}}$  &  $0.07_{\small{({\pm 0.000})}}$ & $0.11_{\small{({\pm 0.002})}}$ & $0.78_{\small{({\pm 0.001})}}$ & $\textbf{0.80}_{\small{({\pm 0.002})}}$ \\
256agents & $0.07_{\small{({\pm 0.000})}}$ & $0.03_{\small{({\pm 0.011})}}$ & $0.47_{\small{({\pm 0.001})}}$ & $\textbf{0.50}_{\small{({\pm 0.001})}}$   &  $0.07_{\small{({\pm 0.000})}}$ & $0.10_{\small{({\pm 0.011})}}$ & $0.48_{\small{({\pm 0.000})}}$ & $\textbf{0.62}_{\small{({\pm 0.000})}}$  \\
512agents &  $0.04_{\small{({\pm 0.000})}}$ & $0.02_{\small{({\pm 0.011})}}$ & $0.16_{\small{({\pm 0.000})}}$ & $\textbf{0.30}_{\small{({\pm 0.000})}}$  & $0.04_{\small{({\pm 0.000})}}$ & $0.04_{\small{({\pm 0.000})}}$ & $0.22_{\small{({\pm 0.000})}}$ & $\textbf{0.24}_{\small{({\pm 0.000})}}$  \\
1024agents & $0.04_{\small{({\pm 0.000})}}$  & $0.02_{\small{({\pm 0.000})}}$ & $0.16_{\small{({\pm 0.000})}}$ & $\textbf{0.30}_{\small{({\pm 0.000})}}$  & $0.04_{\small{({\pm 0.000})}}$  & $0.04_{\small{({\pm 0.000})}}$ & $0.21_{\small{({\pm 0.000})}}$ & $\textbf{0.22}_{\small{({\pm 0.000})}}$   \\ \bottomrule
\end{tabular}%
\caption{Ablation study of training agent population comparing 16 and 32 agents. Performance evaluation after 100 epochs of training. Each
value is reported as the mean ± standard deviation of reach ratio over 10 episodes and 10 seeds}
\end{table*}

\paragraph{Benchmarks.} In this section, we empirically evaluate and analyze SrSv in the widely adopted cooperative multi-agent benchmarks, including the StarCraftII Multi-agent Challenge (SMAC) \cite{samvelyan2019starcraft} with discrete action space and Multi-agent MuJoCo (MA-MuJoCo) \cite{de2020deep} with continuous action space. Besides, we introduced the DubinsCar \cite{zhang2023neural, zhang2024gcbf+} benchmark, which is a fundamental task in mobile robotics, to test the SrSV scalability in large-scale scenarios. To facilitate understanding, we explain the modeling and training target of the DubinsCar benchmark.

For DubinsCar, the local observation of agent \( i \) is given by:
\begin{equation}
    o^i = [p_{x^i}, p_{y^i}, \theta^i, v^i]^{\top}
    \label{eq:dubinscar_obs}
\end{equation}
where \([p_{x^i}, p_{y^i}]^{\top}\) is the position of the agent, \(\theta^i\) is the heading, and \(v^i\) is the speed. The state variables include each agent's local observation, target, and obstacle coordinates. Crucially, the target and obstacle positions are randomly reset upon each episode. The action of agent \( i \) is defined as:
\begin{equation}
    a^i = [\omega^i, \dot{v}^i]^{\top}
\end{equation}
where $\omega^i$ is the angular velocity and $\dot{v}^i$ is the longitudinal acceleration for agent $i$. The dynamics function for agent $i$ is given by:
\begin{equation}
    \dot{o}^i = [v^i \cos(\theta^i), v^i \sin(\theta_i), \omega_i, \dot{v}_i]^{\top} 
\end{equation}

We model the reward function for DubinsCar navigation similar to the settings in \cite{zhang2024gcbf+}, incorporating components for nominal control, goal achievement, and collision avoidance. The specific hyperparameter settings are adapted from \cite{zhang2024gcbf+} to ensure consistency and effectiveness.

\paragraph{Baselines.} We compare SrSV with three baselines: the value decomposition-based method MAPPO, as well as two advantage decomposition-based methods, A2PO and MAT, serving as baselines for sequential update and sequential rollout scheme, respectively. To ensure optimal performance, we use the same hyper-parameters for the baseline algorithms as stated in their original papers. For our methods, we adopt the same hyper-parameter tuning process.

\paragraph{Metrics.} Without loss of generality, we use the \textit{win rate} as the primary metric for SMAC and define it as $\frac{N_w}{N_t}$, where \(N_w\) is the number of wins and \(N_t\) is the total number of games played. We also use the \textit{dead ratio} metric, defined as $\frac{N_d}{N_{total}}$, where \(N_d\) is the number of agents that die during an episode, and \(N_{total}\) is the total number of agents.

For the MA-MuJoCo, we use \textit{the average step reward} metric, defined as $\frac{1}{N_s} \sum_{t=1}^{N_s} r_t$, where \(N_s\) is the total number of steps taken and \(r_t\) is the reward received at step \(t\).

For the DubinsCar, we align with the settings in \cite{zhang2024gcbf+} and evaluate performance using the \textit{reach ratio} metric, defined as $\frac{N_r}{N_c}$, where \(N_r\) is the number of cars that successfully reach the target location within the defined time, and \(N_c\) is the total number of cars.

\subsection{Performance on Cooperative MARL Benchmarks}
We evaluate the algorithms in 5 maps of SMAC with various difficulties, 2 tasks of 1 scenario in MA-MuJoCo, and the DubinsCar scenario with 8-1024 agents and 8 obstacles. Results in Tab. \ref{tab:train_100_300}, Tab. \ref{tab:train_efficiency}, Fig. \ref{fig:training_performance}, Fig. \ref{fig:training_dead_ratio} demonstrate SrSv's superior performance in both training efficiency and equilibrium convergence, while significantly outperforming in all baselines when scaling it to large-scale systems. 


In particular, as shown in Fig. \ref{fig:training_performance}, whether for homogeneous tasks, such as $\texttt{8m vs 9m}$ or for heterogeneous tasks, like $\texttt{MMM2}$ and multi-agent MuJoCo agents, which feature continuous action spaces, SrSv leads to significantly higher training efficiency without compromising convergence performance. More importantly, while SrSv only shows slight improvements in the win rate metric compared to MAT for SMAC tasks (Fig. \ref{fig:training_performance}), it converges to a substantially better equilibrium. As shown in Fig. \ref{fig:training_dead_ratio}, SrSv consistently achieves a lower cost of dead ratio metric in the training phase across all SMAC tasks, indicating that its individual value estimation leads to more effective agent behavior and better individual performance. 

Besides, we adopted the shared parameter mechanism for all the algorithms to verify the model's scalability. As shown in Tab. \ref{tab:train_100_300}, we directly transfer the trained models from the small-scale (8 agents) DubinsCar system to evaluation scenarios with varying population sizes (ranging from 8 to 1024 agents). We demonstrate that SrSv exhibits significantly better scalability than A2PO and MAPPO, whether in the early, or late stages of training. Although in the late stages of training, MAT shows comparable model scalability for small-scale tasks, once scaled up to larger tasks, such as with 1024 agents, MAT's scalability is significantly lower than that of SrSv.

Tab. \ref{tab:train_efficiency} further quantitatively validates SrSv's training efficiency on SMAC tasks using established metrics from \cite{mai2022sample}, including steps to first reach $x$\% performance threshold (SRT) and average training time (ATT). On average across $\texttt{3s5z}$ scenario with 4 random seeds, SrSv requires only 1.12M timesteps (59m) to reach 50\% win rate, significantly outperforming MAPPO (1.30M, 1h), A2PO (1.62M, 1h28m), and MAT (2.00M, 1h42m). Notably, SrSv is the only algorithm besides MAT to achieve 100\% win rate on these SMAC tasks, doing so in considerably less time (3.56M vs 4.92M timesteps).

\begin{table*}[htbp]
\centering
\begin{tabular}{c|ccccc}
\toprule
\multirow{2}{*}{Algo} & 25\% & 50\% & 75\% & 100\% & Total \\
& SRT (ATT) & SRT (ATT) & SRT (ATT) & SRT (ATT) & SRT (ATT)  \\
\midrule
MAPPO & 0.98M (45m) & 1.30M (1h) & 4.34M (3h18m) & - & 5M (3h49m) \\
A2PO & 0.82M (45m) & 1.62M (1h28m) & 2.74M (2h29m) & - & 5M (4h32m) \\
MAT & 1.44M (1h13m) & 2.00M (1h42m) & 2.88M (2h26m) & 4.92M (4h12m) & 5M (4h15m) \\
SrSv & 0.96M (50m) & 1.12M (59m) & 1.52M (1h20m) & 3.56M (3h7m) & 5M (4h23m) \\
\bottomrule
\end{tabular}
\begin{tablenotes}
\small
\item Each cell contains SRT (ATT) values. -: not reach $x$\% win rate during the training phase. M: million timesteps, h: hours, m: minutes.
\end{tablenotes}
\caption{Training efficiency comparison using SRT/ATT metrics for $\texttt{3s5z}$ task}
\label{tab:train_efficiency}
\end{table*}

\subsection{Ablations of SrSv Using Alternative Architectures}

To further demonstrate the necessity of using transformer-based modules for SrSv framework and to explore the performance of alternative architectures, we performed a detailed ablation study include:

\textbf{Non-Attention Methods:} \textit{(1) SrSv-mlp:} This variant replaces the self-attention module with MLPs. All residual connections, layer normalization, and information fusion mechanisms from the SrSv architecture were retained. \textit{(2) SrSv-gnn:} For this variant, we substituted the self-attention module with Graph Neural Networks (GNNs). We adapted the original environment to a graph-based version, with each agent being equally connected to all other agents.

\textbf{Non-Transformer Methods:} \textit{(1) SrSv-enc:} Here, we utilized only the encoder component, excluding the auto-regressive mechanism typically present in the decoder. \textit{(2) SrSv-dec:} In this setup, we used only the decoder component of the transformer, maintaining the auto-regressive property while removing the encoder component, with each agent seeing only its own observation. 

As illustrated in Fig. \ref{fig:architecture_ablation}, taking the \texttt{3s5z} scenario as an example, both the SrSv-mlp and SrSv-gnn variants demonstrate the ability to eventually reach the similar performance of SrSv when maintaining the core architectural features, such as residual connections, layer normalization, and information fusion. Nevertheless, these variants exhibit increased training variance and reduced initial learning efficiency compared to SrSv. Meanwhile, directly employing an encoder or decoder without integrating the transformer architecture of SrSv led to performance degradation. SrSv-dec, in particular, showed a more pronounced decline in win rate metric than SrSv-enc, indicating that observation sharing and embedding among agents are valuable.

\section{Discussion}
This section studies how SrSv affects training efficiency and scalability performance. 
\begin{figure}[t]
  \centering
  \includegraphics[width=\columnwidth]{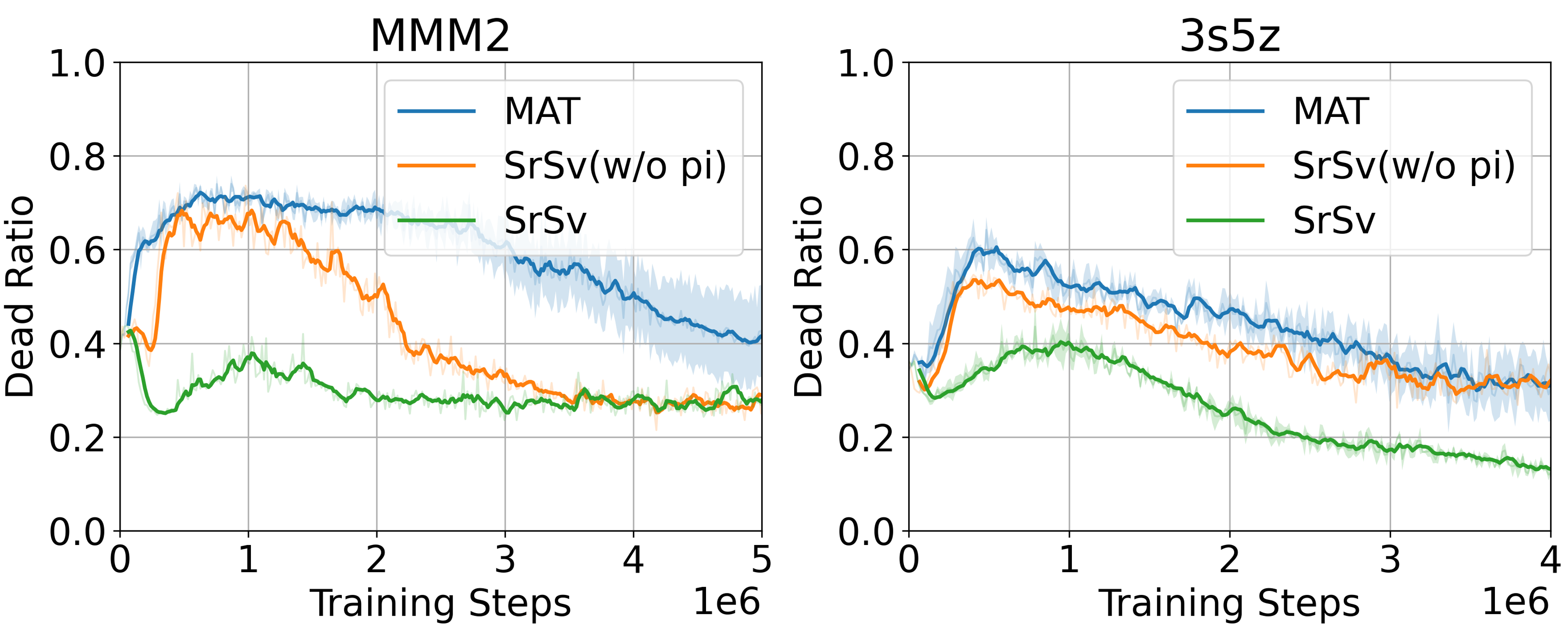} 
  \caption{Ablation study between SrSv and  SrSv (w/o \(\pi\)).}
  \label{fig:ablation}
\end{figure}

\subsection{Is the Training Efficiency Related to $\bm{a}_t^{1:i-1}$ or $\pi_t^{i:n}$?}
In Eq. (\ref{value_function}) and Eq. (\ref{eq:v_pi}), the value function estimation used by SrSv incorporates the specific actions $\bm{a}_t^{1:i-1}$ executed by predecessor agents and the policy distribution $\pi_t^{i:n}$ for successor agents based on the current global policy $\pi_t$. Although the necessity of $\bm{a}_t^{1:i-1}$ has been thoroughly explored in HAPPO and A2PO, its application in sequence value estimation has not been studied. Therefore, to validate the impact of $\bm{a}_t^{1:i-1}$ and $\pi_t^{i:n}$ solely on training efficiency, we designed a variant of SrSv called SrSv (w/o \(\pi\)). Instead of using $V^{\pi _t^{i:n}}(\widehat{\bm{o}_{t}^i}, \bm{a}_t^{1:i-1}, \underset{\bm{a}_t^{i:n}}{\operatorname{argmax}} \ \pi_t^{i:n})$ for individual agent value estimation,  SrSv (w/o \(\pi\)) uses $V^{\pi _t^{i:n}}(\widehat{\bm{o}_{t}^i}, \bm{a}_t^{1:i-1})$ as follows:
\begin{equation}
\label{eq:v_a}
V^{\pi _t^{i:n}}(\widehat{\bm{o}_{t}^i}, \bm{a}_t^{1:i-1}) = \sum_{j=1}^{\textcolor{blue}{i}} w_{i,j} \cdot f_\theta(\widehat{\bm{o}_{t}^j}, e^{j-1})
\end{equation}

As shown in Fig. \ref{fig:ablation}, SrSv (w/o \(\pi\)) exhibits its training efficiency between that of the complete SrSv and MAT, indicating that both \(\pi\) and \(a\) are beneficial for individual value estimation for cooperative multi-agent tasks. Moreover, since $V^{\pi _t^{i:n}}(\widehat{\bm{o}_{t}^i}, \bm{a}_t^{1:i-1}, \underset{\bm{a}_t^{i:n}}{\operatorname{argmax}} \ \pi_t^{i:n})$ aggregates the policy information of all agents, we do not need to consider the impact of decision order on value estimation.

\subsection{Is Scalability Related to the Training Population?}

Furthermore, to better investigate the impact of different numbers of agents during the training phase on scalability, we conducted additional training in the DubinsCar environment using 16 and 32 agents as the number of training agents, training for 100 epochs under the same settings of 8 agents. Then, we transferred the models to handle the test DubinsCar systems with 8 to 1024 agents. 

The experimental results are shown in Tab. \ref{tab:full_population}. Overall, within the tested population sets, as the number of agents during the training phase increases, the overall training complexity also rises. However, SrSv's high training efficiency and scalability capabilities are not affected by the number of training agents, consistently demonstrating significantly better performance than other baselines. Under the same 100 training epochs, the scalability of SrSv declines with an increase in the training population. This trend is also observed in the performance of other baselines.

\section{Conclusion and Future Work}

In this paper, we investigate the potential of leveraging sequential value estimation for sequential decision-making. In particular, a novel paradigm named SrSv, which synergizes individual-centric value estimation from sequential update methods with a sequential rollout strategy, is introduced. It aimed at enhancing the applicability of existing cooperative MARL algorithms in large-scale real-world systems.

We highlight the advantages of the SrSv paradigm from two key perspectives: training efficiency and scalability, comparing it to value decomposition-based and advantage decomposition-based methods. Specifically, through comparative experiments in SMAC and MA-MuJoCo benchmarks, we demonstrate that SrSv significantly improves training efficiency without compromising convergence performance. More importantly, by training in a small-scale DubinsCar system and then transferring to a large scale with 1024 agents, we further affirm the superior scalability of SrSv.

In the future, we will explore how to leverage the advantages of SrSv in capturing inter-agent decision correlations to broaden its application within safe RL, enabling efficient learning of safety-constrained objectives and accelerating the real-world implementation of MARL in industrial scenarios.

\section{Acknowledgments}
This work was supported in part by the National Natural Science Foundation of China under Grants 62103371, 52161135201, the Natural Science Foundation of Zhejiang Province under Grant LZ23F030009, and Alibaba DAMO Academy.

\bibliography{SrSv}
\end{document}